%% file: main.tex
\documentclass[conference]{IEEEtran}

\usepackage{graphicx}
\usepackage[caption=false]{subfig}
\usepackage{amsmath,amssymb}
\usepackage[vlined,algoruled,commentsnumbered]{algorithm2e}
\usepackage{listings}
\usepackage{booktabs}
\usepackage{csquotes}
\usepackage{url}
\usepackage[cleanup={},process=all]{pstool}
\usepackage{anyfontsize}
\usepackage{coverpage}

\title{\LARGE \bf
Sequential path planning for a formation of mobile robots with split and merge
}

\author{M. Estefan\'ia Pereyra$^{1}$, R. Gast\'on Aragu\'as$^{1}$,  and Miroslav Kulich$^{2}$%
\thanks{$^{1}$M. Estefan\'ia Pereyra and R. Gast\'on Aragu\'as are with Research Centre in Informatics for Engineering, National Technological University, C\'ordoba Regional Faculty, Argentina
{\tt\small \{garaguas, mepereyra\}@frc.utn.edu.ar}}%
\thanks{$^{2}$ Miroslav Kulich is with Czech Institute of Informatics, Robotics, and Cybernetics, Czech Technical University in Prague, Prague, Czech Republic
{\tt\small kulich@cvut.cz}}%
}

\begin{document}

\mauthor{M. Estefan\'ia Pereyra, R. Gast\'on Aragu\'as, and Miroslav Kulich}
\published{{\it IEEE Latin American Conference on Computational Intelligence (LA-CCI)}. Arequipa. New Jersey, 2017. ISBN 978-1-5386-3734-0.}
\DOI{10.1109/LA-CCI.2017.8285722}
\original{https://ieeexplore.ieee.org/document/8285722}
\coverpage
\twocolumn
\maketitle

\begin{abstract}
An algorithm for robot formation path planning is presented in this paper.
Given a map of the working environment, the algorithm finds a path for a formation taking into account possible split of the formation and its consecutive merge.
The key part of the solution works on a graph and sequentially employs an extended version of Dijkstra's graph-based algorithm for multiple robots.
It is thus deterministic, complete, computationally inexpensive, and finds a solution for a fixed source node to other node in the graph.
Moreover, the presented solution is general enough to be incorporated into high-level tasks like cooperative surveillance and it can benefit from state-of-the-art formation motion planning approaches, which can be used for evaluation of edges of an input graph.
The performed experimental results demonstrate behavior of the method in complex environments for formations consisting of tens of robots.
\end{abstract}

\input{src/intro}

\input{src/problem}
\input{src/dijkstra}
\input{src/experiments}
\input{src/conclusion}

\section*{Acknowledgments}
The work of M. Estefan\'ia Pereyra and R. Gast\'on Aragu\'as has been supported by the ``Multirrotores Aut\'onomos para Aplicaciones en Ambientes Exteriores'' project, U.T.N. PID UTI4534.
The work of Miroslav Kulich has been supported by the European Union's Horizon 2020 research and innovation programme under grant agreement No. 688117.



\bibliographystyle{IEEEtran}
\bibliography{main}

\end{document}

%% file: src/intro.tex
\section{Introduction}
\label{sec:intro}
Recent advances in mobile robotics and deployment of robotic systems in many practical applications have lead to intensive research of multi-robot systems and robot formations as their special case.
One of the most studied topic deals with trajectory/path planning of a formation, known as multi-robot path planning (MPP) in an environment with obstacles, i.e. the problem, how to find a continuous collision-free motion of the formation through a known environment from a current configuration to a given final configuration.
Besides optimization of some parameters of the solution (e.g., path distance, energy consumption, mission time), a shape and size of the formation is constrained and violation of these constrains is penalized.

Approaches to motion and path planning for multi-robot systems and robot formations can be classified into several categories.
Behavior-based algorithms~\cite{Pereira2008, Zhong2015} are decentralized and reactive, i.e. each robot is controlled individually, using only local information about its neighborhood.
Robot's behavior is typically composed of several simple behaviors (e.g., separation, alignment, and cohesion in~\cite{reynolds1999steering}), which describe basic actions.
These approaches are easy to implement and applicable to large swarms.
They, on the other hand, fail in finding a plan in complex environments and do not guarantee precise formation control.
To deal with the first problem, several heuristic search based algorithms were introduced based on particle swarm optimization~\cite{Bai2009}, genetic algorithms~\cite{Qu2013509} or ant colony optimization~\cite{Asl2014}.
Nevertheless, precise formation shape can not be still maintained.

In contrast, centralized approaches consider a formation as a single body and plan trajectories in a high-dimensional composite configuration space (CCS).
Exact solutions are complete, but their complexity is exponential in the dimension of CCS and therefore, methods based on sampling CSS were introduced.
For example, a probabilistic road map with sampling strategies designed especially for multi-robot systems that enable fast coverage of the configuration space was presented in~\cite{Clark2005}, while a generalization of rapidly exploring random trees (RRT) to a graph structure is introduced in~\cite{Kala13}.

Another research stream considers a leader-follower architecture.
Besides other approaches, which compute leader's trajectory and find trajectories of followers relative to this trajectory~\cite{Barfoot2004, Chen2009} or coordinate motion of robots on a preplanned paths~\cite{olmi2008coordination, Liu2011}, a big class of algorithms employs a concept of artificial potential fields.
As classical potential fields~\cite{Zhang2010} tend to find a local optimum, Garrido et al.~\cite{Garrido2011} employed the Voronoi Fast Marching method, which propagates a wave over a viscosity map for a leader.
Trajectories of followers are then dynamically computed to keep desired nominal inter-robot distances using Fast Marching (FM) with incorporated potentials reflecting leader's path, obstacles and positions of other robots.
The method produces paths with minimal Euclidean lengths and avoids local minimum, but generated paths are not smooth and go too close to obstacles.
This can be solved by Fast Marching Square (FM$^2$)~\cite{Gomez2013}, which modifies wave expansion by incorporating velocity maps.
Moreover, FM$^2$ manages uncertainties in robot's positions, sensor noise, and moving obstacles.
This was recently applied for formations of unmanned surface vehicles (ships) allowing to model their dynamic behavior~\cite{Liu2015}.
Application of the Frenet-Serret frame to control orientation of a formation enabled path planning for formations of unmanned aerial vehicles in 3D environments~\cite{Alvarez2015a}.

The works mentioned above do not explicitly address splitting and merging of a formation during movement, although this is possible with some approaches.
While centralized exact methods are computationally demanding and thus practically inapplicable for larger formations, random sampling are more promising.
For example, the authors in~\cite{Saska2014deployment} present combination of RRT and particle swarm optimization for cooperative surveillance and demonstrate splitting of a formation in a simplified scenario with a single obstacle.
A reactive obstacle avoidance with added rules for split and merge are introduced in~\cite{Ogren2004}, while extension of flocking behavior~\cite{reynolds1999steering} with game-theoretic based reconfiguration is presented in~\cite{dasgupta2013robust}.

The proposed approach can be seen as a part of a general hierarchical planning algorithm, which consists of a global path planner and a local motion planner.
While the the global planner searches for a topological path of a formation and thus generates primarily movement directions, the motion planner determines (based on a path found by the global planner) motion for particular robots in a formation.
This combination prevents the whole planner to be trapped at a local minimum and enables to compute feasible trajectories fast.
Similar approach is described in~\cite{Lin2012}, where the global planner constructs a partial Voronoi diagram on the fly and a memetic evolution algorithm is employed for motion generation along this diagram.
The key contribution of the paper lies in design of a novel algorithm for global path planning, which extents well known Dijkstra's algorithm to be applicable for MPP and which considers possible split and merge of a formation.
This is in contrast with~\cite{Lin2012}, which is primarily focused on a solution of local motion planning, while the global planner is simplified to assume a formation as a single point robot.
On the other hand, motion planning is not addressed in the presented paper as some of the aforementioned approaches can be used for it.
The rest of the paper is organized as follows.
The general overview of the approach is described in Section~\ref{sec:problem}, while the proposed algorithm for robot formations is introduced in Section~\ref{sec:dijkstra}.
In Section~\ref{sec:experiments} we present and discuss some experimental results.
Finally, Section~\ref{sec:conclusion} is dedicated to concluding remarks and future directions of the research.

%% file: src/problem.tex
\section{Multi-robot path planning}
\label{sec:problem}
The planning problem for a fleet of mobile robots, or MPP, can be generally understood as a search of a continuous sequence of feasible fleet configurations from a start configuration to a given goal configuration.
A feasible configuration for a fleet is such configuration that robots in the fleet collide neither with the surrounding environment nor collide with each other.
In many scenarios we don't ask for an arbitrary sequence of configurations (trajectory), an optimal path with respect to some criteria (e.g., a mission time) is required instead.
Moreover, additional constrains to fleet's geometry may be applied.
For example, robots may be requested to form and keep a specified shape or lattice, or to move close together in order to ensure visibility or communication to their neighbors.

In this work, we investigate the problem of MPP, from a common start point to a common end point of a known map.
Each robot path is obtained by minimizing a cost function, which is proportional to the path length, the number of robots moving in the same path, and inversely proportional to the path width (i.e. space between obstacles).
The formation can split if necessary but it is forced to merge as soon as possible, what means that each robot arriving to a common point should wait to all other robots of the formation that include this point on their paths.
Consequently, the cost of the complete solution is given by the most expensive individual path.
A valid solution of MPP is that which avoids collisions with obstacles in the map and collisions between robots.
A collision between robots occurs when two or more robots share the same part (edge) of the path but in the opposite direction.
To avoid collisions between robots a set of constrains for resulting paths are imposed, as described in Section~\ref{sec:pathsconstrains}.
To avoid collisions with obstacles, we construct a connected graph representing topology of collision free space of the working area as described in the next section and perform MPP on this graph.

\subsection{General framework}
\begin{figure}[!htb]
  \centering
  \subfloat[][]{ \label{fig:vor-a}
  \includegraphics[width=0.5\columnwidth]{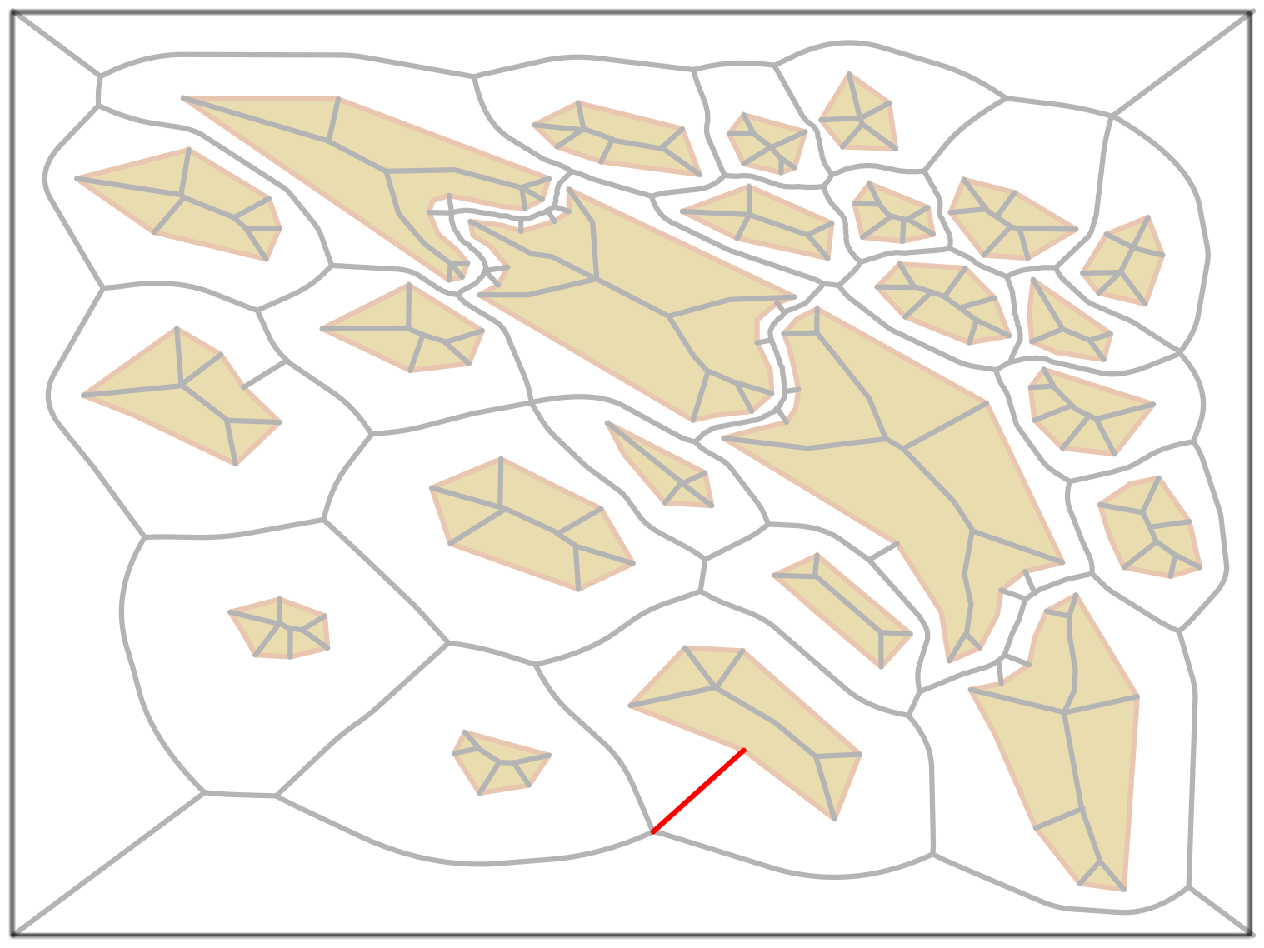}}
  \subfloat[][]{ \label{fig:vor-b}
  \includegraphics[width=0.5\columnwidth]{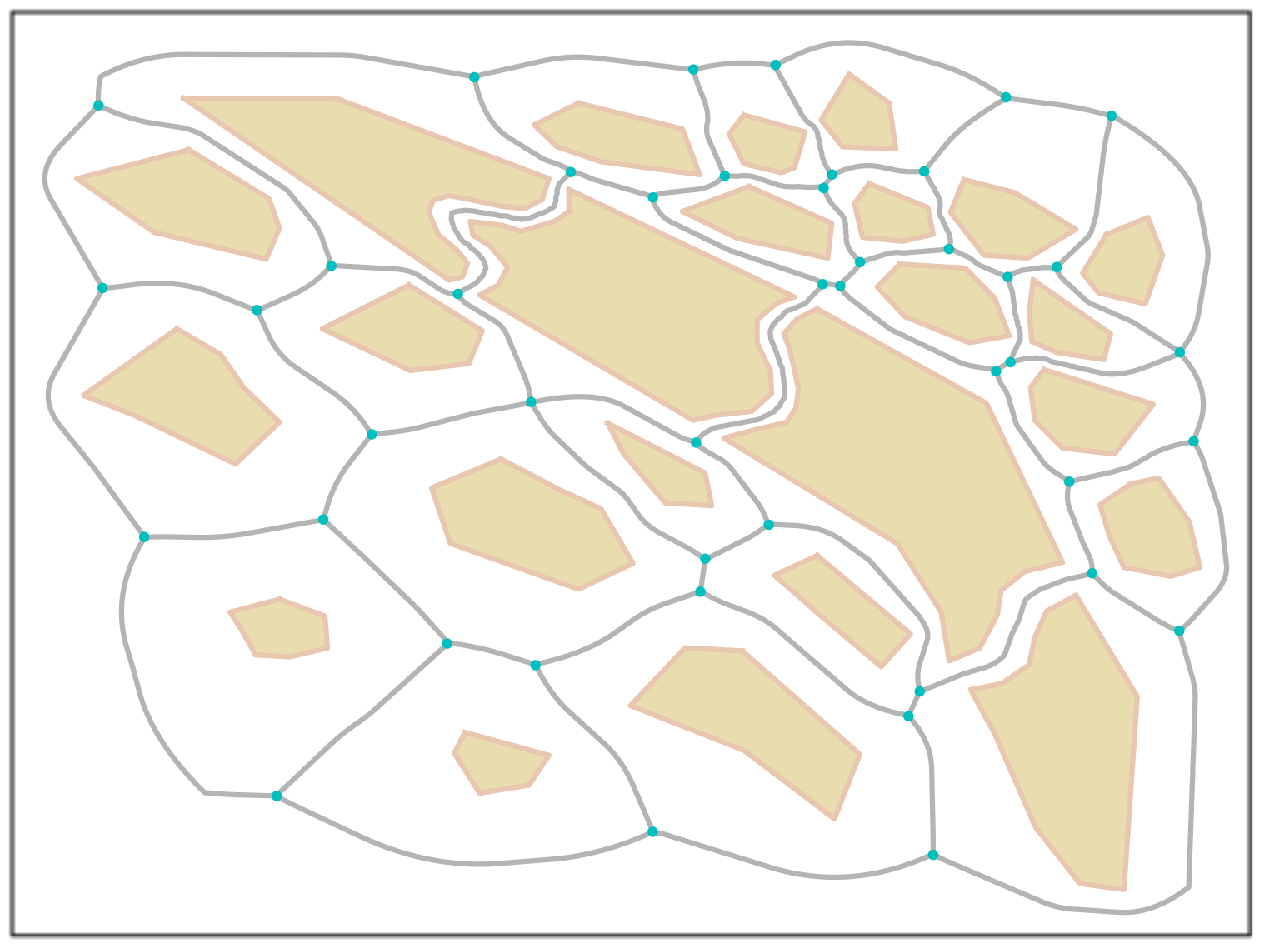}}
  \caption[Graph construction.]{Graph construction.  \subref{fig:vor-a} A Voronoi diagram (light blue) of obstacles (beige) and the rectangular border\footnotemark.  The red line represents a tail to be filtered.  \subref{fig:vor-b} The same after filtering of edges and vertices.}
  \label{fig:vor}
\end{figure}

Given a polygonal representation of the environment (Fig.~\ref{fig:vor}\subref{fig:vor-a}), start and goal positions, and the number of robots $R$, the proposed solution depicted in Algorithm~\ref{alg:framework} works as follows.
\footnotetext{We use implementation of a Voronoi Diagram (VD) from the Boost Polygon Library (\url{http://www.boost.org/doc/libs/1\_60\_0/libs/polygon/doc/voronoi\_diagram.htm}). The image was generated directly from the output of the library. Note that a Voronoi Diagram for polygons contains generally parabolic segments which are approximated with line segments in the image.}
It starts with the construction of a connected graph $\cal G(V,E)$ of vertex $\cal V$ and edges $\cal E$ (line~\ref{es1}), which is done in three steps.
\LinesNumbered
\begin{algorithm}[ht]
Construct a connected graph $\cal G(V,E)$\nllabel{es1}\\
Evaluate each edge $e\in \cal E$\nllabel{es2}\\
For a given number of robots $R$, compute a shortest path in $\cal G(V,E)$ from the start node $Q_{\rm ini} \in \cal V$ to the goal node $ Q_{\rm goal} \in \cal V$ for each robot\nllabel{es3}\\
Generate motion along the found shortest path to the given goal node\nllabel{es4}\\
\caption{The general MPP framework.}
\label{alg:framework}
\end{algorithm}
A Voronoi diagram of all obstacles in the environment is generated (Fig.~\ref{fig:vor}\subref{fig:vor-a}), resulting in a set of polygons rounding the obstacles, and edges which are inside obstacles or which are connected to some obstacle vertex.
These edges are then removed together with nodes and edges forming tails, i.e. components of the graph which can not be a part of any shortest path except paths originating or ending in these components (one of these components is highlighted in red in  Fig.~\ref{fig:vor}\subref{fig:vor-a}).
The graph after this removal is shown in  Fig.~\ref{fig:vor}\subref{fig:vor-b}.
Finally, if the goal or final positions are not in the graph, they are added into it.
All remaining edges of the graph are evaluated furthermore (line~\ref{es2}).
The cost of particular edges is a vector $\boldsymbol{c}=(c_1,c_2,\dots c_R)$, where $c_r$ is the cost of a formation of $r$ robots to traverse the edge.

\begin{figure}[!htb]
  \centering
  \includegraphics[width=0.95\columnwidth]{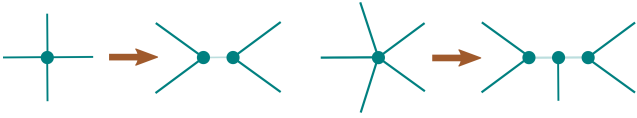}
  \label{fig:split}
  \caption{Substitution of a node with a degree ($\rho$) higher than 3. A node with $\rho=4$ is substituted with two nodes, which are connected with an edge with cost equal to zero (left). Similarly, node with  $\rho=5$ is substituted with three nodes (right). Edges with the zero cost are shown in the light color.}
\end{figure}

The cost of an edge can be either determined as the time needed to traverse the edge with some motion planning algorithm (e.g.~\cite{Gomez2013}) in simulation or it can be approximated based on the length of the edge and the distance of the edge to the nearest obstacle.
Some control coefficient can also be used to modify the cost for $r<R$ and thus to manage the formation splits. 
For simplicity the algorithm processes all nodes in the graph assuming that they have a degree less or equal to three,  $\rho\leq 3$, that is, they have two or three connected edges.  
Nodes with $\rho > 3$ are substituted by an equivalent set of nodes-edges that meet this constraint.
Some examples of this substitution are shown in Fig.~\ref{fig:split}. 
The proposed algorithm for formations is run next (line~\ref{es3}), which for a given start node computes shortest paths to the goal node in $\cal G(V,E)$ and all possible sizes of the formation.
Each of the computed paths is a sequence of nodes $p = \{v_0=Q_{\rm ini}, v_1, v_2, \dots, v_n=Q_{\rm goal} \}$.
The algorithm is described in Section~\ref{sec:dijkstra} in details.
A complete motion of a formation is generated finally, given the path $p$ (line~\ref{es4}).
Trajectories of robots in the formation are determined in this step so that relative positions of robots are computed  with respect to geometrical constrains on the formation and to avoid nearby obstacles. 
Again, some motion planning algorithm can be employed to plan a motion between each two consecutive nodes of $p$.

\subsection{Path properties and constrains}
\label{sec:pathsconstrains}
Given a connected graph $\cal G=(V,E)$ representing the working environment, with a set of vertices ${\cal V} = \{v_1, v_2, \dots, v_i\}$ and edges ${\cal E} = \{e_1, e_2, \dots, e_j\}$, and a fleet of robots  ${\cal R} = \{r_1, r_2, \dots, r_R\}$, the corresponding collision-free paths for the fleet in the graph are denoted as ${\cal P} = \{p_1, p_2,\dots, p_R\}$, with $p_i: \mathbb Z^+ \to \cal V$.
Moreover, a path of the individual robot $r_i$ is a sequence of vertices $p_i = \{v_{i_1}, v_{i_2}, \dots v_{i_k}\}$ such that $(v_{i_j}, v_{i_{j+1}})$ is an edge of the graph.

The feasibility of a path $p_i$ is conditioned upon the following constrains: 1) initially, all the robots in the fleet are in the start position: $p_i(0) = Q_{\rm ini}, \,\forall p_i \in \cal P$.
2) there exists a state in the path $k_{\rm min}\in \mathbb Z^+$ such that $p_i(k_{\rm min}) = Q_{\rm goal}$, meaning that the robot $r_i$ reaches the goal on the shortest possible path. 
3) any two paths from ${\cal P}$ have not to be in a collision, i.e. given any pair of states $m,l \in \left<0, k_{\rm min}\right>$, two paths $p_i, p_j$ are in collision if $(p_i(m), p_i(m+1)) = p_j(l+1), p_j(l))$.
4) given two paths $p_i, p_j$ and two states $m,l \in \left<0, k_{\rm min}\right>$, if $p_i(m) = p_j(l)$ then $m = l \gets max(m,l)$, i.e. the robot that arrives first to a vertex waits for the second one, in order to keep the formation joint as much time as possible.

%% file: src/dijkstra.tex
\section{Sequential path planning for formations}
\label{sec:dijkstra}
The proposed algorithm for MPP is explained in details in this section.
Basically, the algorithm finds paths for a multi-robot team running sequentially an extended version of the well known Dijkstra's algorithm for one robot on a tailored graph. 
The standard Dijkstra's algorithm finds the cheapest paths together with their costs from the given source node $Q_{\rm ini}$ to all other nodes in the graph $\cal G$. 
The algorithm stores for each node $v\in {\cal G}$ the minimum cost $C_{v_{\rm min}}$ to reach this node and the predecessor $v_{\rm prev}$ from where the node is reached.
The shortest path (in terms of cost) for the given node $v$ can be then easily determined by walking consecutively over predecessors starting in $v$.
In the initialization stage, costs $C_v$ of all the nodes are set to infinity and their predecessor $v_{\rm prev}$ to a fictive node $None$, what means that the shortest path has not been found yet.
The only exception is the start node, whose cost value is set to $0$.
The start node is then put into a priority queue, which is sorted according to $C_{v_i}$.
The algorithm then consecutively takes the nodes from the priority queue and for the current node $u$, their neighbors $v_i$ are processed.
For each neighbor $v_i$, the total cost $C_{v_i}$ of arriving to it from $u$ is computed.
If $v_i$ is reached for the first time by the algorithm, it is added into the priority queue with its total cost $C_{v_i}$ corresponding of arrive to it from $u$, and $u$ is assigned as its predecessor, $v_{\rm prev} = u$.
If it is not the case, the computed cost $C_{v_i}$ is compared with the cost storaged in the node and, if it is smaller, the storaged cost is updated to $C_{v_i}$, $v_{\rm prev}$ is set to $u$, and $v_i$ with its new cost is added to the priority queue.

For the multi-robot case, the cost of an edge can vary with the size of the formation, and thus, all possible combinations given by the number of robots traveling through the node must be considered.
This means that a node can not be processed at once, because it is not guaranteed that all needed information is available until all the robots are processed.
Based on the standard Dijkstra's algorithm, path planning for a multi-robot formation can be computed in a semi-decoupled way, considering split and rejoin of ẗhe formation.
The path planning is performed sequentially for all robots, revisiting each node accordingly to the number of robots that will travel trough it.

Given a formation of $R$ robots, a connected graph $\cal G(V,E)$, and a vector of costs ${\boldsymbol c}^e = \left( c^e_1, c^e_2, \dots, c^e_R\right)$ for each edge  $e$, where $c^e_r$ is the cost paid for traversing $e$ with $r \in \left<1,R\right>$ robots, the sequential algorithm for a formation is depicted in Algorithm~\ref{alg:sequential}.
Similarly to the standard Dijkstra's algorithm, the proposed sequential algorithm starts with initialization of data structures (lines~\ref{ald:init1}~--~\ref{ald:endinit}).
Here, for each node of the graph, all predecessors are set to $None$ and the corresponding costs are initialized to infinity, except the cost to reach $Q_{\rm ini}$, which is set to $0$.
After that, all the nodes are pushed into the queue $\cal H$ (line~\ref{alg:addtoqueue}).
\begin{algorithm}
  \caption{Sequential algorithm for MPP\label{alg:sequential}}
  \KwIn{$\cal G(V,E)$ - connected graph\\
  $R$ - number of robots\\
  $Q_{\rm ini}$ - start node\\
  $Q_{\rm goal}$ - goal node}
  \KwOut{$\cal P$ - set of paths}
  \BlankLine
  \hrule
  \BlankLine
  \ForEach{number of robots $r \in \{1, \dots, R\}$}{\label{ald:robots}
    \ForEach{node $v \in {\cal G(V,E)}$}{\label{ald:init1}
    \If{$v =  Q_{\rm ini}$}{
      $C_{\rm v} \gets 0$\\
      $v_{\rm prev} \gets None$\\
    }
    \Else{
      $C_{\rm v} \gets \infty$\\
      $v_{\rm prev} \gets None$
    }
    ${\cal H}.add(\left<C_{\rm v} ,v_{\rm prev}\right>)$ \label{alg:addtoqueue}
  }\label{ald:endinit}
  \While{not goal}{\label{ald:loop}
      $\left<C_{\rm u}, u_{\rm prev}\right> = {\cal H}.pop()$\\
      \ForEach{neighbour $v$ of $u$}{
        \If{$v \notin \cal N$}{
          $r_{\rm uv} \gets robots\_in\_segment()$\label{ald:upcost1}\\
          \If{$(C_{\rm u} + c^{r_{\rm uv}}_{\rm uv}) < C_{\rm v} $}{
            $\left<C_{\rm v}, v_{\rm prev}\right> \gets \left<(C_{\rm u} + c^{r_{\rm uv}}_{\rm uv}), u\right>$\label{ald:upcost2}
          }
          ${\cal H}.add(\left<C_{\rm v},v_{\rm prev}\right>)$
        }\label{ald:opt_ini}
        ${\cal N}.add(u)$\label{ald:opt_end}
      }
    }
    $p_{\rm r} \gets compute\_path({\cal P}_{\rm r-1})$\label{ald:path}\\
    ${\cal P}_{\rm r-1}.add(p_{\rm r})$
 }
\end{algorithm}
The algorithm then loops through all nodes in the queue and processes them in a similar way to the original Dijkstra's algorithm (line~\ref{ald:loop}).
When the map is processed completely, the path for the first robot is generated at line~\ref{ald:path}, and added to the set of paths at the next line.
Next, the path for the second robot in the formation is planned, updating costs of those edges in the graph that are a part of the previously computed robot path (lines~\ref{ald:upcost1}~--~\ref{ald:upcost2}).
The algorithm ends whenever the paths ${\cal P}_R = \{p_1, p_2,\dots,p_R\}$ for all $R$ robots are computed. 

\subsection{Path optimization}
\begin{figure}[hbt]
  \centering
  \subfloat[][]{\label{fig:example-a}
  \includegraphics[width=0.45\columnwidth]{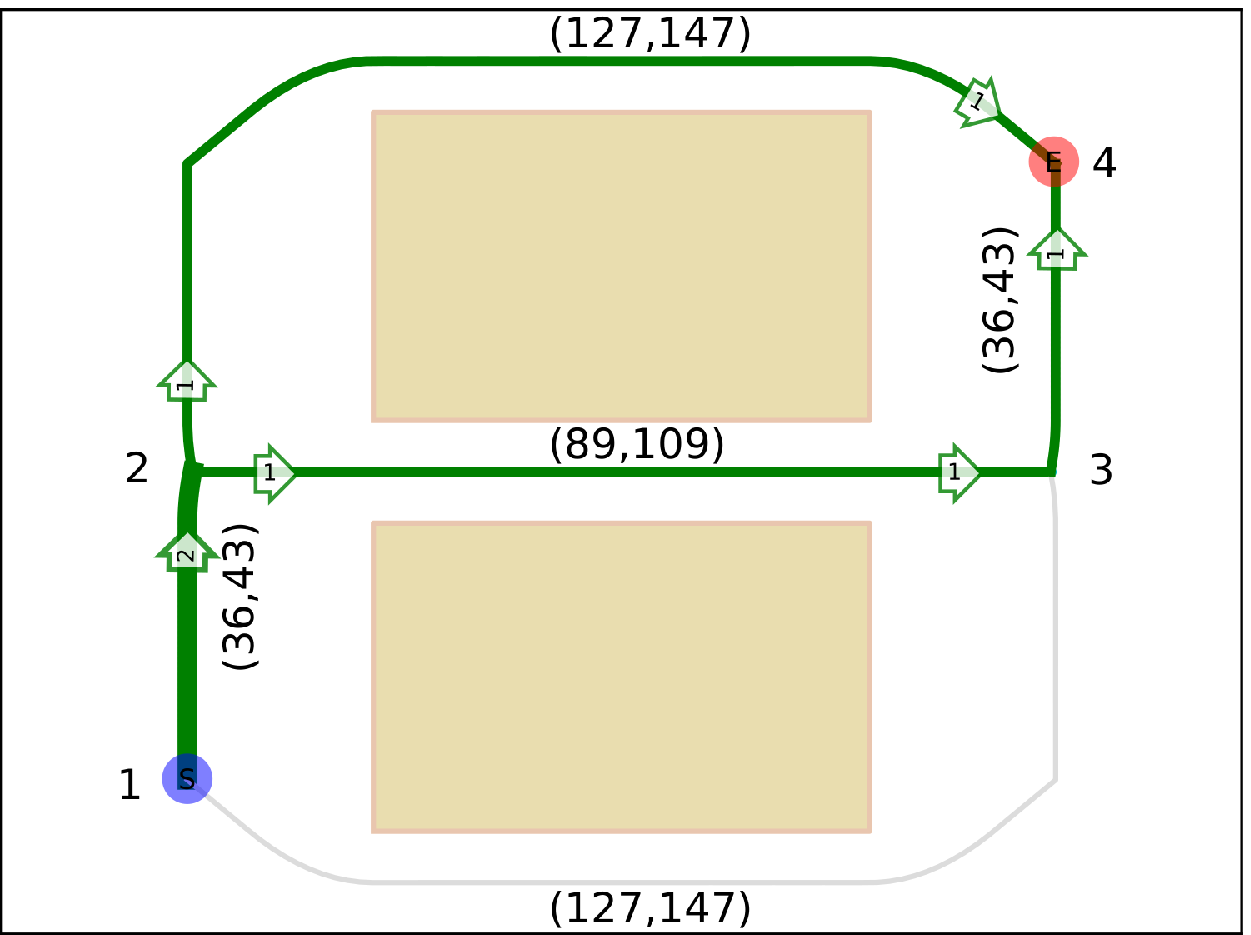}}
  \hfill
  \subfloat[][]{\label{fig:example-b}
  \includegraphics[width=0.45\columnwidth]{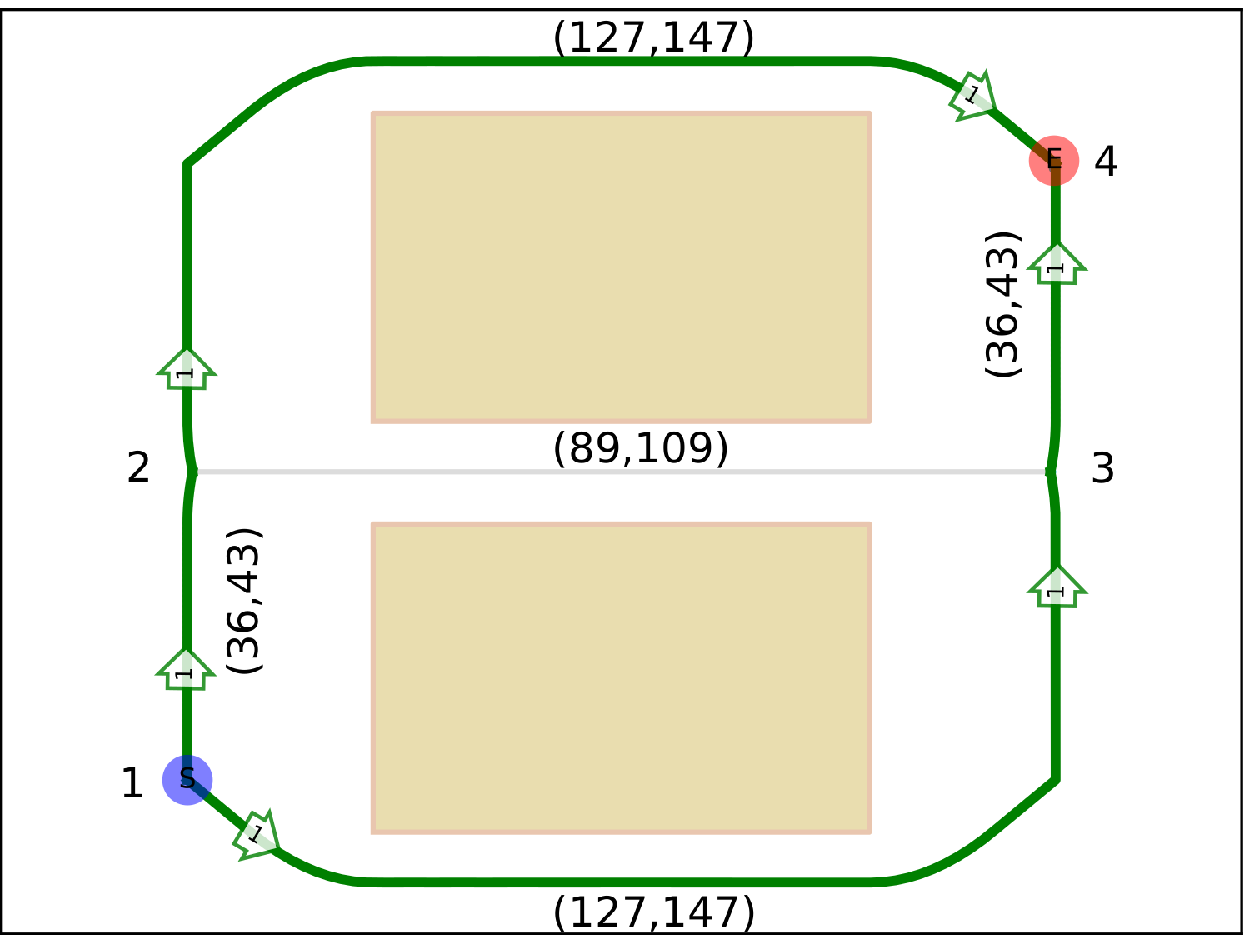}}

  \caption[Graph construction.]{Graph construction.
    \subref{fig:example-a} Two robots planning from $Q_{\rm ini}$ (blue) to $Q_{\rm goal}$ (red), without optimization.
    \subref{fig:example-b} Optimized two robots planning.
  }
  \label{fig:example}
  \vspace{-1em}
\end{figure}

The paths ${\cal P}_R = \{p_1, p_2,\dots,p_R\}$ for $R$ robots computed as explained in the previous section are optimal if the paths ${\cal P}_{R-1}$ calculated for $R-1$ robots are a part of the optimal solution for $R$ robots, i.e ${\cal P}_{R-1} \subset {\cal P}_{R}$.
For example, in Fig.~\ref{fig:example}\subref{fig:example-a} the paths obtained for two robots are shown.
The paths are represented by green lines and the arrows over each edge indicate the number of robots traversing it.
The cost vector  ${\boldsymbol c}^e = \left( c^e_1, c^e_2\right)$ for each edge  $e$ is shown nearby the edge.
In this case, the paths found by the algorithm from $Q_{\rm ini}=1$ to $Q_{\rm goal}=4$ are $p_1 = \{1, 2, 3, 4\}$, with cost $C_1 = 168$ and $p_2 = \{1, 2, 4\}$, with cost $C_2 = 170$.   
The optimal paths for two robots between these nodes are nevertheless $p_1 = \{1, 3, 4\}$ and $p_2 = \{1, 2, 4\}$, with costs $C_1 = C_2 = 163$.
These two optimal paths are shown in Fig.~\ref{fig:example}\subref{fig:example-b}.  
This problem can be partially solved by means of an optimization process applied in the second stage of the algorithm. 
For each new calculated path $p_{\rm r}$, the function $compute\_path()$ (line~\ref{ald:path}) performs the following optimization process.
Once a new path is calculated, all the previous paths are recalculated sequentially, beginning from the first one.
If a recalculated path results in a lower cost of the formation, the original path is replaced by this new one.
The optimization process performed by this function is detailed in Algorithm~\ref{alg:optimization}. 
\begin{algorithm}
  \caption{Paths optimization\label{alg:optimization}}
  \KwIn{
    $\cal G(V,E)$\\
    ${\cal P}_{\rm r}$ - partial set of paths\\
    $r$ - partial number of robots\\
    $Q_{\rm ini}$ - start node\\
    $Q_{\rm goal}$ - goal node}
    \KwOut{${\cal P}_{\rm r}$ - set of optimal paths}
    \BlankLine
    \hrule
    \BlankLine

    \ForEach{path $i \in \{1, \dots, r\}$}{ \label{alg:foreachpath}
    ${\cal P}_{\rm aux} \gets {\cal P}_{\rm r}$ \label{alg:storeP}\\
    $C_{\rm prev} \gets max\_cost(\cal P_{\rm aux})$\label{alg:cost}\\
    $p_{\rm i} \gets {\cal P_{\rm aux}}.pop()$\label{alg:popp}\\
    $p_{\rm new} = sequential\_search(\cal P_{\rm aux})$\label{alg:newpath}\\
    ${\cal P_{\rm aux}}.add(p_{\rm new})$\label{alg:addpnew}\\
    $C_{\rm new} \gets max\_cost(\cal P_{\rm aux})$\label{alg:newcost}\\
    \If{$C_{\rm new} < C_{\rm prev}$}{ \label{alg:comparecosts}
      ${\cal P}_{\rm r} \gets {\cal P}_{\rm aux}$ \label{alg:newpaths}
    }
  }
\end{algorithm}
Given a set of paths ${\cal P}_{\rm r} = \{p_1, p_2, \cdots, p_{\rm r}\}$, the algorithm runs over all previously calculated paths (line~\ref{alg:foreachpath}).
A copy of the current set of paths ${\cal P}_r$ is stored to be processed (line~\ref{alg:storeP}), and the maximum cost of the set is computed at line~\ref{alg:cost}.
A subset ${\cal P}_{\rm r-1} \subset {\cal P}_{\rm r} \setminus p_{\rm i}: i \in \left<1, \rm r-1\right>$ is extracted then (line~\ref{alg:popp}) and a new path is computed (line~\ref{alg:newpath}), in a similar way as in the Algorithm~\ref{alg:sequential} but considering the subset ${\cal P}_{\rm r-1}$ to determine the edges costs.
This new path is added to the subset ${\cal P}_{\rm r-1}$, and the maximal cost of $\{{\cal P}_{\rm r-1}\}\cup p_{\rm new}$ is computed (lines~\ref{alg:addpnew} and \ref{alg:newcost}). 
If the cost is lower than the lod one, the new computed path $p_{\rm new}$ is preserved (lines~\ref{alg:comparecosts} and \ref{alg:newpaths}).

\subsection{Formation control coefficient}
\begin{figure}[hbt]
  \centering
  \includegraphics[width=0.48\columnwidth]{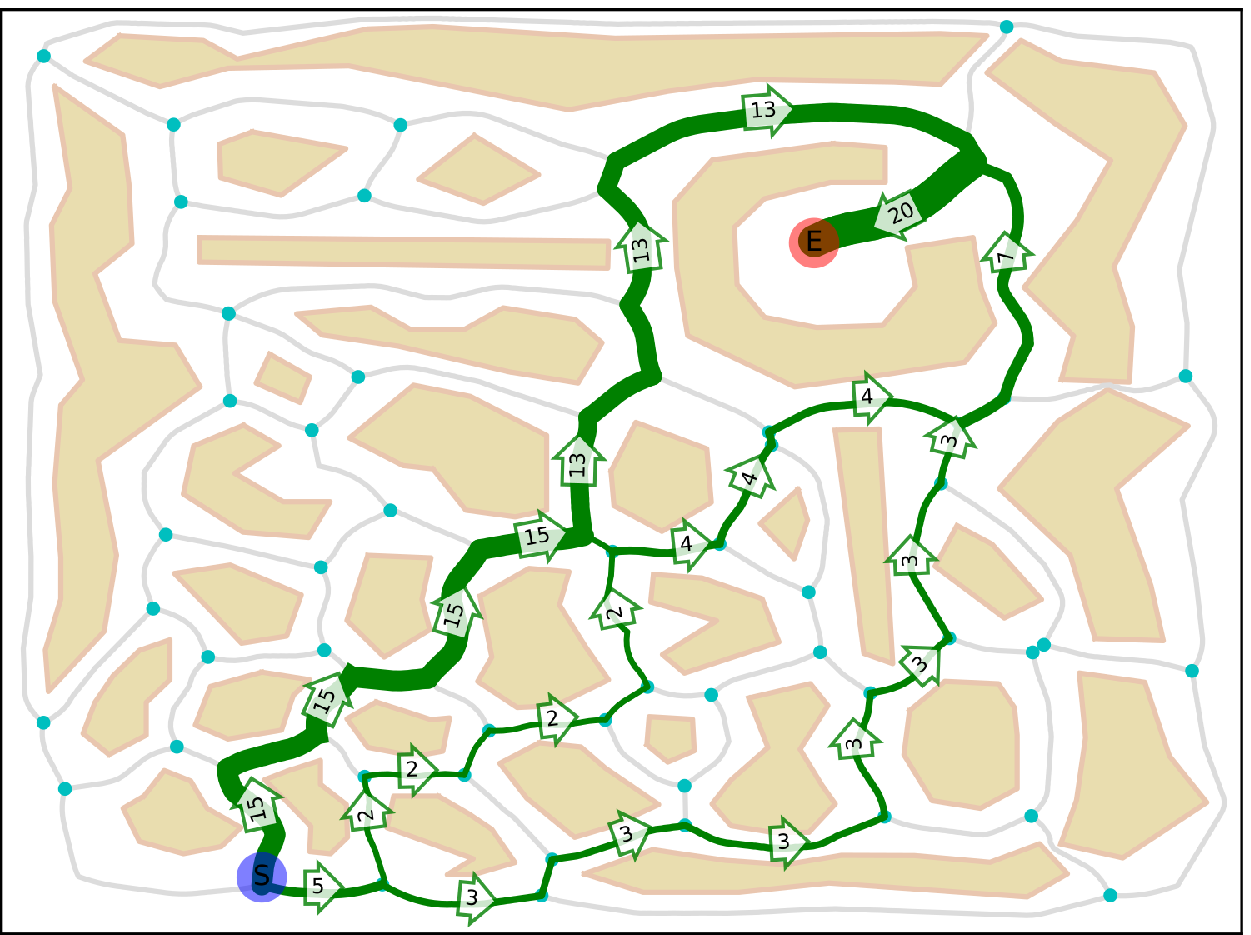}
  \includegraphics[width=0.48\columnwidth]{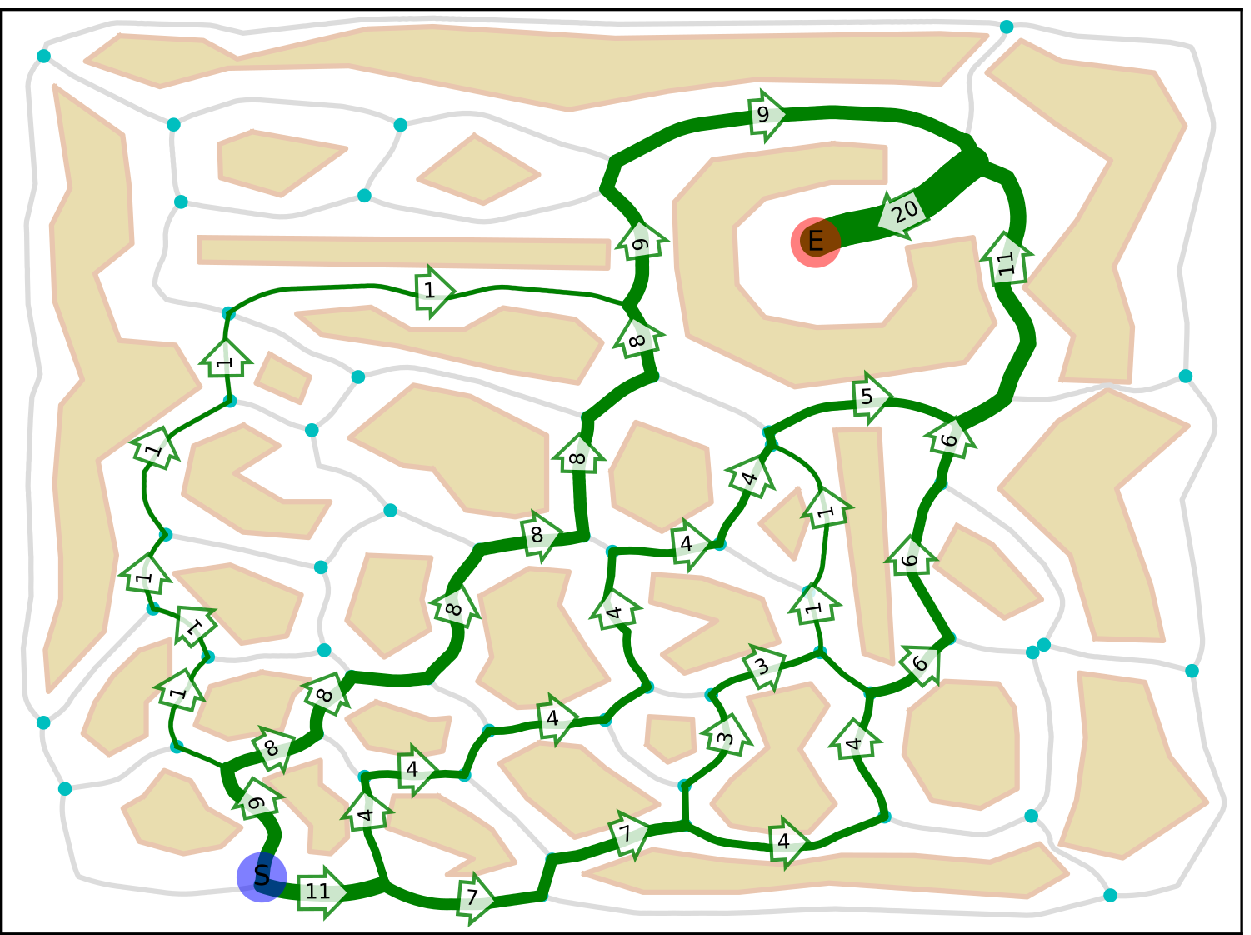}
  \caption{MPP for $20$ robots with various values of the formation control coefficient $k$, with $k=10$ for the left map and $k=100$ for the right one.
  }
  \label{fig:mdense-example}
  \vspace{-1em}
\end{figure}

To keep control over the joint of the formation a coefficient $k$ named \emph{formation control coefficient} is used.
This coefficient modifies the cost of each edge proportionaly to the number of robots. 
The paths computed using different values of the coefficient $k$ are shown in Fig.~\ref{fig:mdense-example}. 
For a low $k$ the formation tends to stay together as shown in the left map ($k=10$), while a high value of $k$ makes the formation to split over all the map and cover big areas, as shown in the right map of Fig.~\ref{fig:mdense-example}.

%% file: src/experiments.tex
\section{Experimental evaluation}
\label{sec:experiments}
\begin{figure*}[!t]
  \centering
  \includegraphics[width=0.65\columnwidth]{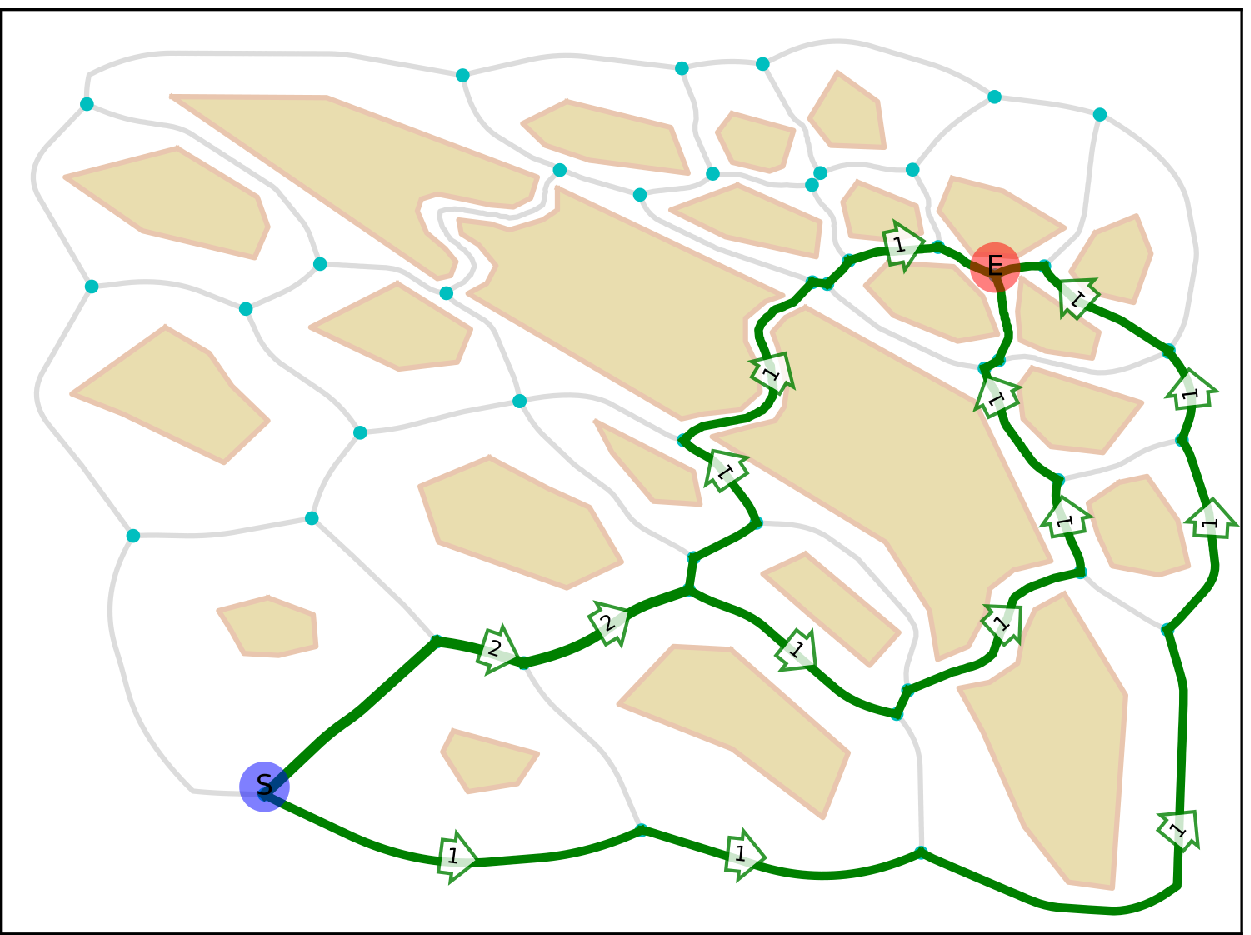}
  \hfill
  \includegraphics[width=0.65\columnwidth]{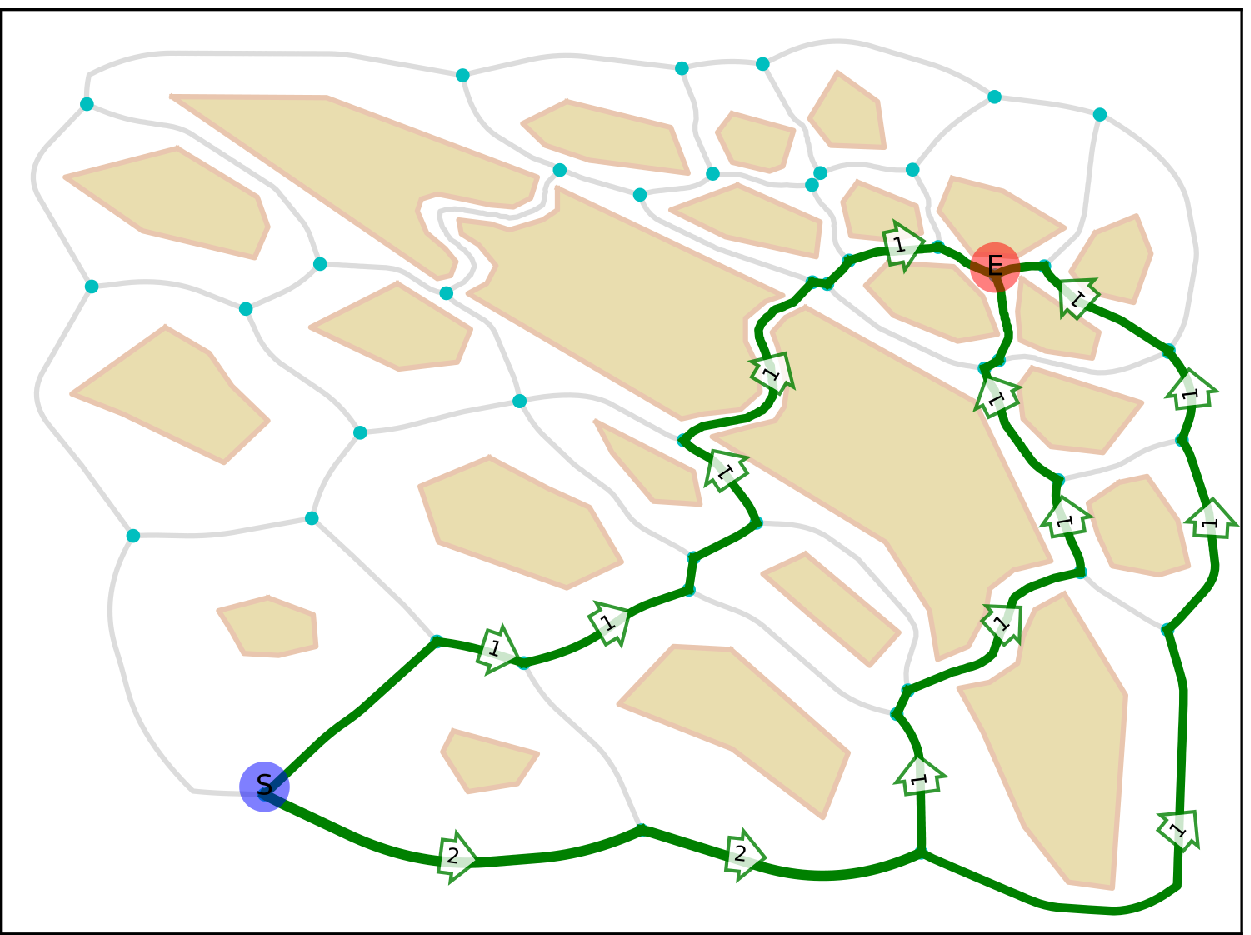}
  \hfill
  \includegraphics[width=0.65\columnwidth]{fig/var_density2_robots3_start21_end29_k1000_exhaustive}
  \caption{MPP for $R=3$. The map in the left shows the optimal paths, with costs $C_{\rm goal} =\{412, 346, 372 \}$, in the map in the middle sowhs the MPP without optimization where the costs are $C_{\rm goal} = \{335, 376, 418\}$, and the map in the right shows the optimized MPP with costs $C_{\rm goal} =\{412, 346, 372 \}$. 
  }
  \label{fig:var_density2-example}
  \vspace{-1em}
\end{figure*}

\begin{figure*}[!t]
  \centering
  \subfloat[][$169$ out of $197$ paths were optimal ($85.78\%$)]{\label{fig:data_analysis_nonopt}
  \includegraphics[width=0.62\columnwidth]{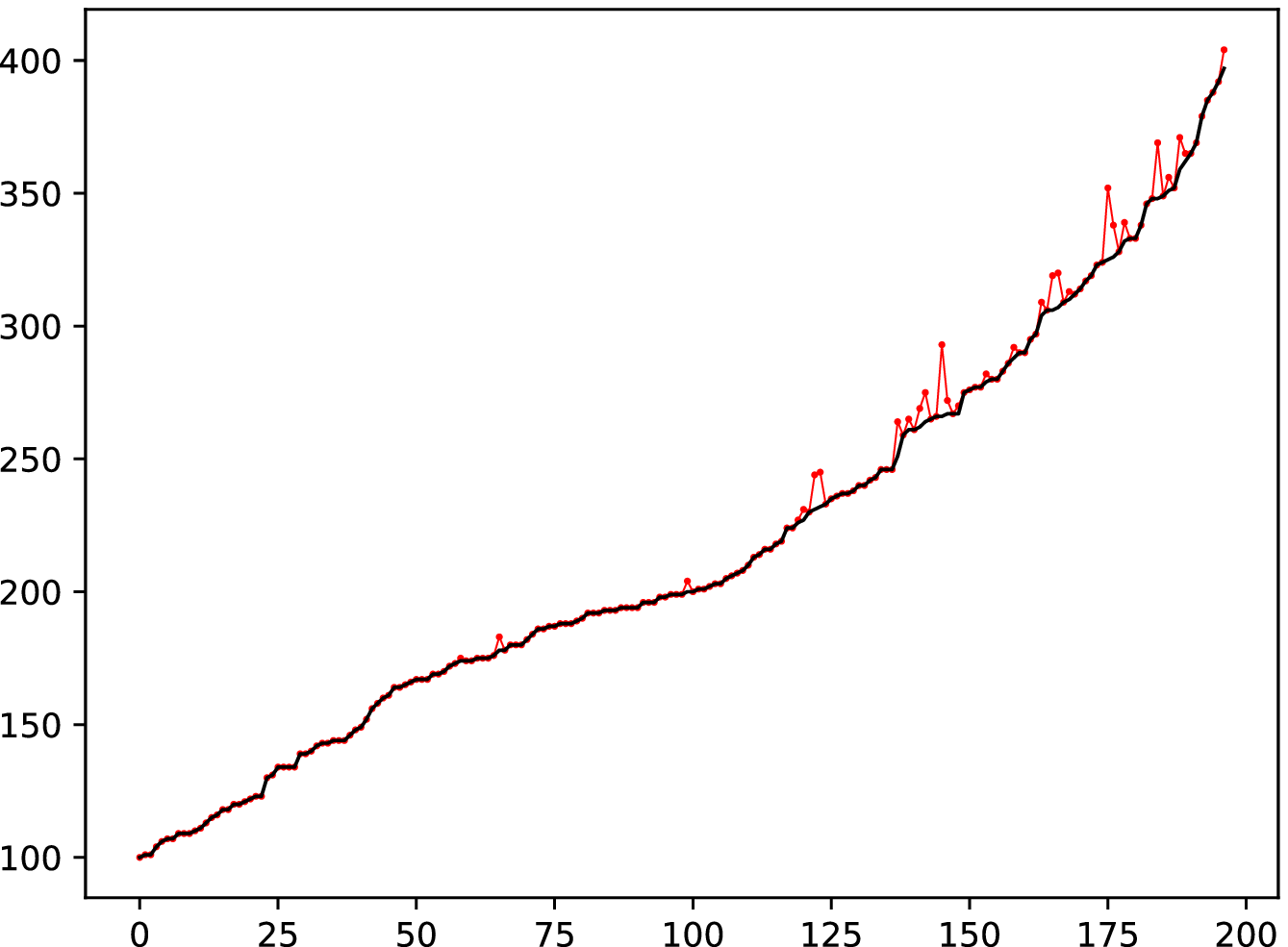}}
  \hfill
  \subfloat[][$181$ out of $197$ paths were optimal ($91.88\%$)]{\label{fig:data_analysis_opt}
  \includegraphics[width=0.62\columnwidth]{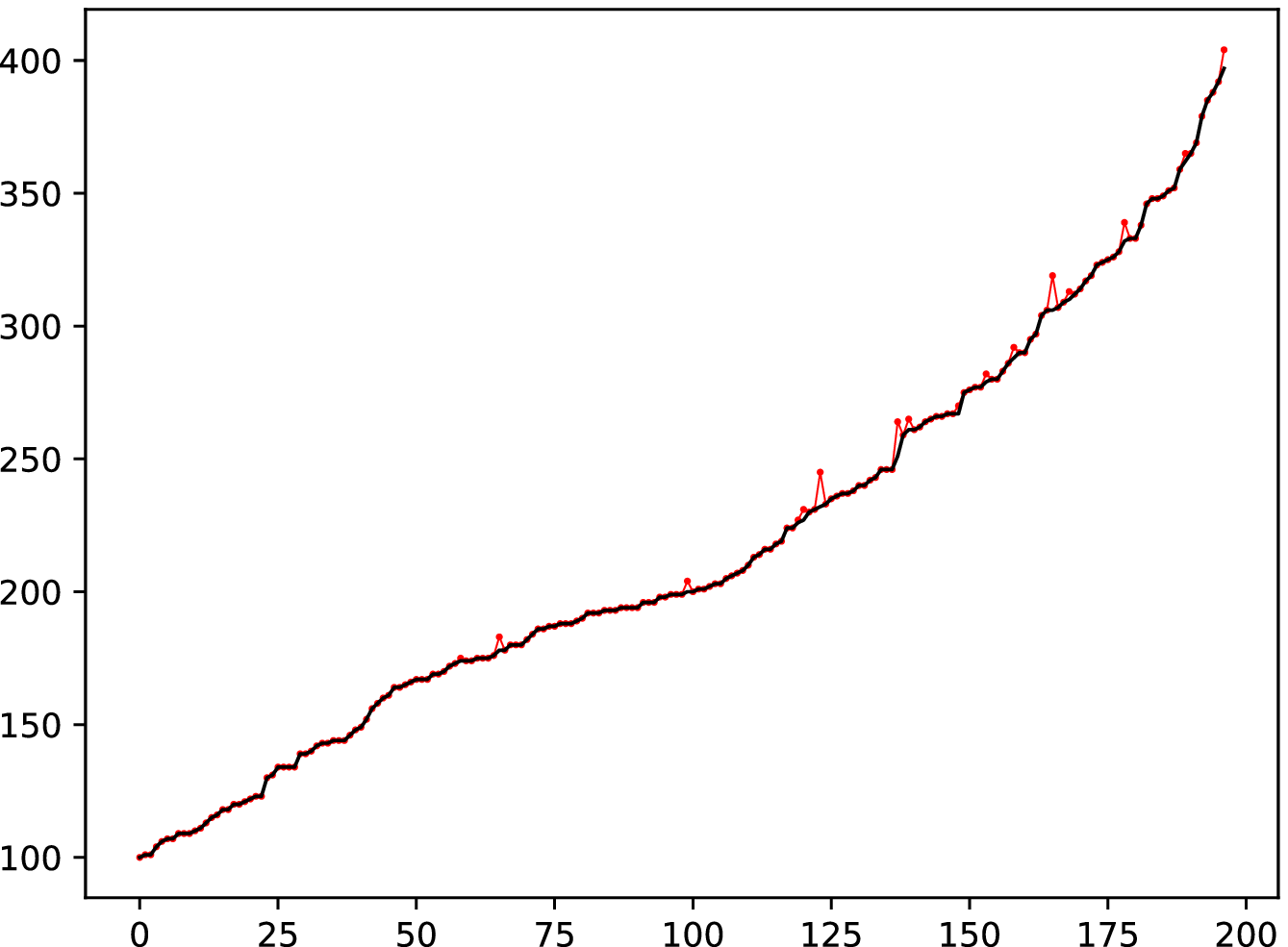}}
  \hfill
  \subfloat[][Optimized ({\tiny {\bf +}}), not optimized ({\tiny$\color{red}{\bullet}$}) gap.]{\label{fig:data_analysis_relative_error}
  \includegraphics[width=0.62\columnwidth]{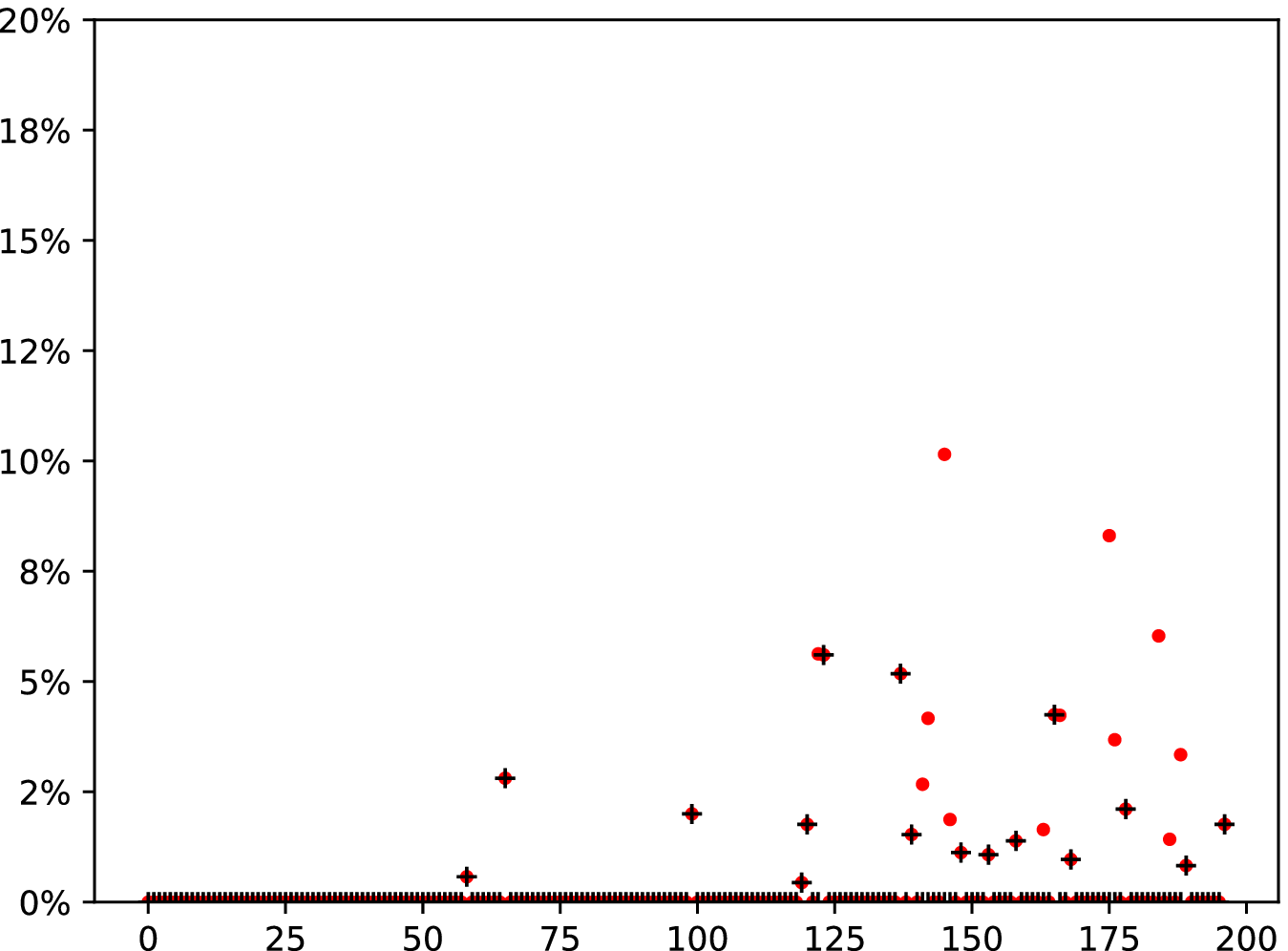}}
  \caption{The first two plots show comparison of optimal (black) and sequential solutions (in red) without optimization in \protect\subref{fig:data_analysis_nonopt} and with optimization in \protect\subref{fig:data_analysis_opt}.
  \emph{x}-axis = experiment number and \emph{y}-axis = cost of the corresponding path.
  \protect\subref{fig:data_analysis_relative_error} is the percentage gap given by the optimized ({\tiny {\bf +}})  and non-optimized ({\tiny$\color{red}{\bullet}$}) cases.
  }
  \label{fig:data_analysis}
\end{figure*}

All the experiments were evaluated under the same conditions, using a notebook with an \texttt{Intel CORE i5} processor and 4GB of RAM, running a Debian GNU/Linux. 
The proposed algorithm has been implemented in Python 2.7.
The maps used in the experiments are \emph{gaps}, \emph{dense}, \emph{staggered\_brick\_wall}, \emph{potholes}, \emph{var\_density}, \emph{var\_density2} and \emph{var\_density3} from \url{http://imr.ciirc.cvut.cz/planning/maps.xml}.
An optimal path for each configuration (map, ${\cal{Q}}_{\rm ini}$, ${\cal{Q}}_{\rm goal}$, $R$,  $k$) is obtained by an exhaustive search method, which is used for comparison.
The exhaustive search has a high computational complexity, which grows exponentially with the number of robots and the number of nodes in the map, and it is thus very difficult to obtain results for more than three robots in maps with more than ten nodes. 
Each test is performed by running the Algorithm~\ref{alg:sequential}, with and without the optimization process given as described in the Algorithm~\ref{alg:optimization}, and both results are compared with optimal results generated by the exhaustive search algorithm.
Fig.~\ref{fig:var_density2-example} shows a typical test, where the optimization process performs a paths correction resulting in the same solution as the one provided by the exhaustive search.
Fig.~\ref{fig:data_analysis} shows a comparison of the optimal paths (black lines) and the paths generated by the proposed algorithm (red lines): without the optimization process in Fig.~\ref{fig:data_analysis_nonopt} and with optimization in Fig.~\ref{fig:data_analysis_opt}.
The plots are sample \emph{vs} cost, where a sample refers to a path returned by the algorithm for each (map, ${\cal{Q}}_{\rm ini}$, ${\cal{Q}}_{\rm goal}$, $R$, $k$) configuration, and cost is the cost of the returned path.
From a total of $\approx2000$ computed paths, about $85\%$ solutions generated without optimization is optimal, while about $92\%$ solutions is optimal after optimization is applied.
Note that only 10\% of samples is drawn in Figs.~\ref{fig:data_analysis_nonopt} and \ref{fig:data_analysis_opt} and the data are sorted by cost ascending for better visualization.
In order to describe the quality of the found solution relative to the optimum, a gap $g = (c - c_{OPT}) / c_{OPT}$ is calculated for each sample, where $c$ is the cost of the found solution and $c_{OPT}$ is the cost of the optimal solution.
Fig.~\ref{fig:data_analysis_relative_error} shows the gap between each sequential solution and the optimal path for both non-optimized and optimized cases.
Table~\ref{tab:seq_times}  presents the runtime of the proposed path planning algorithm for various numbers of robots and various maps with optimization (opt) and without it (nonopt). 
We set $k=1000$, since the process time is barely affected by this value.
The most left-bottom node of the map was chosen as ${\cal Q}_{\rm init}$ while the most right-top node was set to be ${\cal Q}_{\rm goal}$ in all the tests.
Side-by-side comparison with the state-of-the-art methods is not presented due to (seemingly small but) important differences in the problem formulation.
Nevertheless, the computation time of the proposed algorithm including the optimization process appears comparable with results from other authors, while the computation time of the non-optimized version of the algorithm is several times lower.
\begin{table*}[htb]
  \centering
  \setlength\tabcolsep{0.5ex}
  \begin{tabular}{ l  c  p{1ex}rp{1ex}l p{1ex}rp{1ex}l p{1ex}rp{1ex}l p{1ex}rp{1ex}l p{1ex}rp{1ex}l }
    \toprule
    \multicolumn{2}{r}{ } & \multicolumn{20}{c}{Number of robots}  \\
    \cmidrule{3-22}
    \multicolumn{2}{r}{ } & \multicolumn{4}{c}{5} & \multicolumn{4}{c}{10} & \multicolumn{4}{c}{20} & \multicolumn{4}{c}{50} & \multicolumn{4}{c}{100} \\
    \midrule
    maps &(nodes)             && opt~ &-& nonopt && opt~ &-& nonopt && opt~ &-& nonopt && opt~ &-& nonopt && opt~ &-& nonopt \\
    \hline
     {\bf gaps         }&(12)  && 0.002 &-& 0.0004 && 0.013 &-& 0.001 && 0.097 &-& 0.002 && 1.157 &-& 0.011 && 9.166 &-& 0.036\\
     {\bf staggered\_bk}&(33)  && 0.006 &-& 0.001 && 0.045 &-& 0.003 && 0.263 &-& 0.010 && 3.291 &-& 0.042 && 26.780 &-& 0.125\\
     {\bf potholes     }&(43)  && 0.007 &-& 0.002 && 0.047 &-& 0.005 && 0.373 &-& 0.016 && 4.622 &-& 0.067 && 33.429 &-& 0.198\\
     {\bf var\_density2}&(45)  && 0.010 &-& 0.003 && 0.058 &-& 0.006 && 0.310 &-& 0.016 && 4.613 &-& 0.066 && 39.447 &-& 0.202\\
     {\bf var\_density3}&(45)  && 0.012 &-& 0.002 && 0.058 &-& 0.006 && 0.353 &-& 0.017 && 5.495 &-& 0.070 && 39.152 &-& 0.205\\
     {\bf dense        }&(62)  && 0.010 &-& 0.002 && 0.054 &-& 0.007 && 0.391 &-& 0.021 && 5.425 &-& 0.083 && 37.789 &-& 0.233\\
     {\bf var\_density} &(77)  && 0.019 &-& 0.004 && 0.114 &-& 0.012 && 0.609 &-& 0.035 && 10.438 &-& 0.141&& 76.157 &-& 0.409\\
    \bottomrule
  \end{tabular}
\vspace{0.6em}
  \caption{Times in seconds with (\textnormal{opt}) and without (\textnormal{nonopt}) the optimization process, for $5$, $10$, $20$, $50$ and $100$ robots with $k=1000$.}
  \label{tab:seq_times}
\vspace{-3.2em}
\end{table*}

%% file: src/conclusion.tex
\section{Conclusion}
\label{sec:conclusion}
A new decoupled approach for multi-robot path planning with split and merge capability was presented.
The algorithm is deterministic, complete and computationally inexpensive.
The performed tests on various maps show that the algorithm provides the optimal solution in approximately  $84\%$ of cases which can be increased to $92\%$ when a simple optimization procedure is applied.
Solutions that remain suboptimal have nevertheless a low gap, which means that resulting paths are close to the optimum.
Time complexity of the algorithm is very low as shown in the experiments, what makes the algorithm suitable for real-time (re-)planning or its deployment on on-board computers of real robots.
The future work will be focused on improvement of the optimization process.